\pdfoutput=1

\documentclass[11pt]{article}

\usepackage[]{acl}

\usepackage{times}
\usepackage{latexsym}

\usepackage[T1]{fontenc}

\usepackage[utf8]{inputenc}

\usepackage{microtype}

\title{OPERA: Operation-Pivoted Discrete Reasoning over Text}

\author{Yongwei Zhou\textsuperscript{1}, Junwei Bao\textsuperscript{2}\thanks{~~Work was done during the first author’s internship at JD AI mentored by Junwei Bao: baojunwei001@gmail.com.}, Chaoqun Duan\textsuperscript{2}, Haipeng Sun\textsuperscript{2}, Jiahui Liang\textsuperscript{2}, \\
    \bf Yifan Wang\textsuperscript{2}, Jing Zhao\textsuperscript{2}, Youzheng Wu\textsuperscript{2}, Xiaodong He\textsuperscript{2}, Tiejun Zhao\textsuperscript{1}\thanks{~~Corresponding author.} \\
    \textsuperscript{1}Harbin Institute of Technology \;\;  \textsuperscript{2}JD AI Research \\
    ywzhou@hit-mtlab.net \;\;  baojunwei@jd.com  \;\; tjzhao@hit.edu.cn
    }
    
\usepackage{bibentry}

\usepackage[utf8]{inputenc}
\usepackage{subfigure}
\usepackage{booktabs}
\usepackage{amsmath}
\usepackage{amsfonts}
\usepackage{multirow}
\usepackage{color}
\usepackage{enumitem}
\usepackage{microtype}
\usepackage{tabularx}

\usepackage{chngpage}
\usepackage{array}
\usepackage{microtype}
\usepackage{xcolor}
\usepackage{fontenc}
\usepackage{arydshln}
\usepackage{array}
\usepackage{booktabs}

\usepackage{newfloat}
\usepackage{listings}
\usepackage{makecell}
\usepackage{aliascnt}

\usepackage{algorithm}
\usepackage{algorithmic}
\usepackage{hyperref}
\usepackage{graphicx}

\newcommand{\tabincell}[2]{\begin{tabular}{@{}#1@{}}#2\end{tabular}}

\usepackage[utf8]{inputenc}

\usepackage{cleveref}
\crefname{section}{§}{§§}
\Crefname{section}{§}{§§}

\begin{document}
\maketitle

\begin{abstract} 
Machine reading comprehension (MRC) that requires discrete reasoning involving symbolic operations, e.g., addition, sorting, and counting, is a challenging task.
According to this nature, semantic parsing-based methods predict interpretable but complex logical forms.
However, logical form generation is nontrivial and even a little perturbation in a logical form will lead to wrong answers.
To alleviate this issue, multi-predictor -based methods are proposed to directly predict different types of answers and achieve improvements.
However, they ignore the utilization of symbolic operations and encounter a lack of reasoning ability and interpretability.
To inherit the advantages of these two types of methods, we propose \textbf{OPERA}, an 
operation-pivoted discrete reasoning framework, where lightweight symbolic operations (compared with logical forms) as neural modules are utilized to facilitate the reasoning ability and interpretability.
Specifically, operations are first selected and then softly executed to simulate the answer reasoning procedure.
Extensive experiments on both DROP\footnote{\url{https://leaderboard.allenai.org/drop/submissions/public}} and RACENum datasets show the reasoning ability of OPERA.
Moreover, 
further analysis 
verifies its interpretability.
\footnote{Codes is released at \url{https://github.com/JD-AI-Research-NLP/OPERA}.}
\end{abstract}

\section{Introduction}

Machine reading comprehension (MRC) that requires discrete reasoning is a valuable and challenging task \cite{dua2019drop},
especially involving symbolic operations such as addition, sorting, and counting.
The examples in Table \ref{example} illustrate the task.
To answer the question ``\textit{Who threw the longest pass}”, 
it requires a model to choose the person with the longest pass from all the people and corresponding distance pairs based on the given passage.
This task has various application scenarios in the real world, such as analyzing financial reports and sports news.

Existing approaches for this task can be roughly divided into two categories: the semantic-parsing-based
\cite{chen2019neural,gupta2019neural} and the multi-predictor-based methods \cite{dua2019drop,ran2019numnet,hu2019multi,chen2020question,zhou2021evidr}. 
\begin{table}[t]
    \centering
    \scalebox{.9}{
    \begin{tabularx}{0.53\textwidth}{X}
        \hline \hline
        {\textbf{Passage}:
        ~~~Houston would tie the game in the second quarter with kicker Kris Brown getting a \textcolor{blue}{53}-yard and a \textcolor{blue}{24}-yard field goal. Oakland would take the lead in the third quarter with wide receiver Johnnie Lee Higgins catching a \textcolor{blue}{29}-yard touchdown pass from Russell, followed up by an \textcolor{blue}{80}-yard punt return...
        }\\
        \midrule
       \textbf{Question}:~~\textcolor{cyan}{Who threw} the \textcolor{cyan}{longest} pass?\\
        \textbf{Answer}:~~~~Russell\\
        \hline \hline
    \end{tabularx}
    }
    \caption{\label{example} An example of question-answer pair along with a passage from DROP dataset \cite{dua2019drop}. 
    Question words in color indicate the potential operations for reasoning, i.e., 
    \textcolor{cyan}{$\texttt{ARGMAX}$} and \textcolor{cyan}{$\texttt{KEY\_VALUE}$}.}
    \vspace{-0.6cm}
\end{table}%
The former maps the natural language utterances into logical forms and then executes them to derive the answers.
For example, \citet{chen2019neural} propose NeRd. 
It includes a reader to encode the passage and question, and a programmer to generate a logical form for multi-step reasoning.
Intuitively, this method has an advantage in interpretability.
However, semantic parsing over text is nontrivial and even a little perturbation will result in wrong answers, which hinders the MRC performance.

To alleviate the heavy dependence on logical forms in the first category, the latter directly employs multiple predictors to derive different types of answers.
For example, \citet{dua2019drop} and \citet{chen2020question} divide instances of the DROP dataset into several types and design
a model with multi-predictors to deal with different answer types, i.e., question/passage span(s), count, and arithmetic expression.
It is the capability of deriving different types of answers that improves the performance of models.
However, such methods are lack of the necessary components to imitate discrete reasoning, which leads to inadequacy in reasoning ability and interpretability.

To alleviate the shortcomings of the above methods and preserve their advantages, we attempt to summarize reasoning steps into a set of operations and adopt them as the pivot to connect the question and the answer, which makes it possible to perform discrete reasoning.
For example, to answer the question in Table~\ref{example}, it needs two steps: (1) finding all persons and the corresponding distance of touchdown pass, and (2) choosing the one with the longest pass among them.
We attempt to convert them into two operations, \texttt{KEY\_VALUE} and \texttt{ARGMAX}, respectively.
We then use them to produce the answer.
Specifically, we design a set of lightweight symbolic operations (compared with logical forms) to cover all of the questions in the datasets and utilize them as neural modules to facilitate reasoning ability and interpretability.
We denote this method as \textbf{OPERA}, an operation-pivoted discrete reasoning MRC framework.
To utilize the operations, we propose an operation-pivoted reasoning mechanism composed of an operation selector and an operation executor.
Specifically, the operation selector automatically identifies relevant operations based on the input.
To enhance the performance of this sub-mechanism, we further design an auxiliary task to learn the alignment from a question to operations according to a set of heuristics rules.
The operation executor softly integrates the selected operations to perform discrete reasoning over text via an attention mechanism \cite{vaswani2017attention}.

To verify the effectiveness of the proposed method, comprehensive experiments are conducted on both the DROP and RACENum datasets, where RACENum used in this paper is a subset of the RACE dataset following \cite{chen2020question}.
Experimental results indicate that our method outperforms strong baselines and achieves the state-of-the-art on both datasets under the single model setting.
We further analyze the interpretability of OPERA.
Overall, this paper primarily makes the following contributions:

(1) We propose OPERA, an operation-pivoted discrete reasoning MRC framework, improving both the reasoning ability and interpretability.

(2) Extensive experiments on DROP and RACENum dataset demonstrate the reasoning ability of OPERA. 
Moreover, statistic analysis and visualization indicate the interpretability of OPERA.

(3) We systematically design operations and heuristic rules to map questions to operations, aiming to facilitate research on symbolic reasoning.

\begin{table*}[htpb]
  \small
  \centering
  \setlength\tabcolsep{10pt}
  \resizebox{\textwidth}{!}{
\begin{tabular}{m{0.15\textwidth} m{0.47\textwidth} m{0.56\textwidth}}
\hline

 \bf Operations & \bf Description & \bf{Examples}\\ \midrule

\texttt{ADDITION/DIFF} &  Addition or subtraction & \textit{How many more yards was Kris Browns's first field goal over his second?}  \\ \midrule

\texttt{MAX/MIN} & Select the maximum/minimum one from given numbers & \textit{How many yards was the longest field goal in the game?}  \\ \midrule

\texttt{ARGMAX/ARGMIN } & Select key with highest/lowest value from key-value pairs & \textit{Which player had the longest touchdown reception?} \\
\midrule

\texttt{ARGMORE/ARGLESS} & Select key with higher/lower value from two key-value pairs & \textit{Who scored more field goals, David Akers or John Potter?} \\ 
\midrule

\texttt{COUNT} & Count the number of spans & \textit{How many field goals did Kris Brown kick?}\\ \midrule

\texttt{KEY\_VALUE} &Extract key-value pairs & \textit{How many percent of Forth Worth households owned a car?}\\ \midrule
 
\texttt{SPAN} & Select spans from input sequence & \textit{Which team scored the final TD of the game?}\\

  \hline 
  \end{tabular} }
   \caption{\label{Operators} All the operations, descriptions and the corresponding examples.}
\end{table*}

\section{Related Work}

Recently, machine reading comprehension (MRC) methods tend to deal with more practical problems~\cite{yang2018hotpotqa,dua2019drop,zhao-etal-2021-ror-read}, for example, answering complex questions that require discrete reasoning~\cite{dua2019drop} such as arithmetic computing, sorting, and counting.
\label{sec:related_work}
Intuitively, semantic parsing-based methods, which are well explored to deal with discrete reasoning in question answering with structured knowledge graphs~\cite{bao-etal-2016-constraint} or tables, have potential to address the discrete reasoning MRC problem.
Therefore, semantic parsing-based methods for discrete reasoning over text are proposed to firstly convert the unstructured text into a table, and then answer questions over the structured table with a grammar-constrained semantic parser \cite{krishnamurthy2017neural}.
NeRd \cite{chen2019neural} is a generative model that consists of a reader and a programmer, which are responsible for encoding the context into vector representation and generating grammar-constrained logical forms, respectively. NMNs \cite{gupta2019neural} learned to parse compositional questions as executable logical forms. %
However, it only adapts to limited question types matched with a few pre-defined templates. 

Multi-predictor-based methods employ multiple predictors to derive different types of answers. NAQANET \cite{dua2019drop}, a number-aware framework, employed multiple predictors to produce corresponding answer types, including a span, count, and arithmetic expression. Based on NAQANET, MTMSN\cite{hu2019multi} added a negation predictor to solve the negative question and re-ranked arithmetic expression candidates.
To aggregate the relative magnitude relation between two numbers, NumNet~\cite{ran2019numnet} and NumNet+ leveraged a graph convolution network to perform multi-step reasoning over a number graph. 
QDGAT \cite{chen2020question} proposed a question-directed graph attention network for reasoning over a heterogeneous graph composed of entity and number nodes. 
EviDR \cite{zhou2021evidr}, an evidence-emphasized MRC method, performed reasoning over a multi-grained evidence graph.
Compared with these existing methods, our proposed OPERA focuses on bridging the gap from questions to answers with operations and integrating them to 
simulate discrete reasoning.

\section{Approach\label{approach_section}}

\subsection{Task and Model Overview}
\label{sec:task_formulation}
Given a question $Q$ and a passage $P$, MRC that requires discrete reasoning aims to predict an answer $\hat{A}$ with the maximum probability over the candidate space $\Omega$ as follows:
\begin{equation}
    \begin{split}
      \hat{A} =\arg \max_{A \in \Omega}p(A|Q,P)
    \end{split}
\end{equation}
where the answer $\hat{A}$ in this task could not only be span(s) extracted from context but also a number calculated with some numbers in context.
To handle this task, it generally requires not only natural language understanding but also performing discrete reasoning over text, such as comparison, sorting and arithmetic computing.

\begin{figure*}[tb]
	\centering
	\includegraphics[width=6.3in]{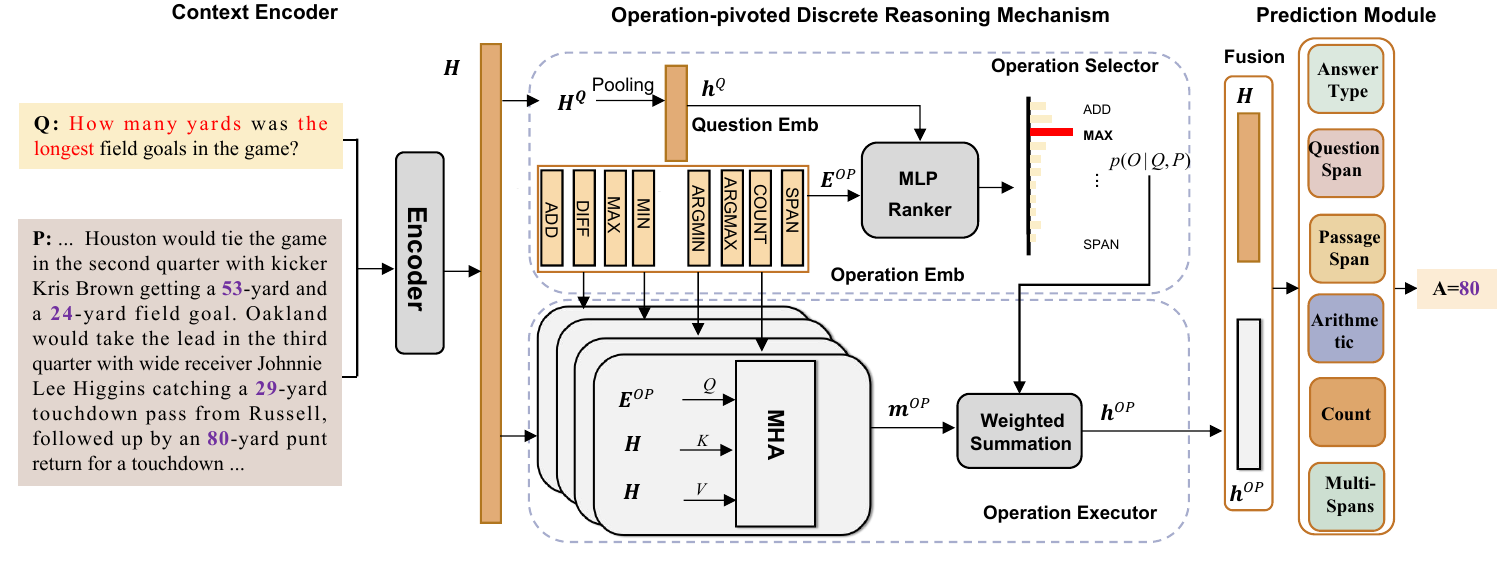}
	\caption{\label{model} The architecture of OPERA. It consists of a context encoder, an operation-pivoted reasoning module, and a prediction module. The prediction module supports five types of answers, including question span, passage span, arithmetic expression, count, and multi-spans. MHA means a multi-head attention mechanism.}
\end{figure*}

To address the aforementioned challenges in this task, we propose OPERA, an operation-pivoted discrete reasoning MRC framework and it is briefly illustrated in Figure \ref{model}.
In our framework, a set of operations $\mathcal{OP}$, defined in Table \ref{Operators}, are introduced to support the modeling of answer probability $p(A|Q, P)$ as follows:
\begin{equation}
    p(A|Q,P) = \!\sum_{O \in \mathcal{OP}}{p(A|Q,P,O) p(O|Q,P)},
\end{equation}
where $O\in\mathcal{OP}$ represents one of the operations. 
Concretely, in our framework, we first design an operation selector $p(O|P, Q)$ for choosing the correct question-related operations.
These selected operations are then \textit{softly} executed over the given context. Eventually, answer predictor $p(A|Q,P,O)$ utilizes the execution result to predict the final answer.

\subsection{Definition of Operations}
\label{sec:Mathematical_Operations}

To imitate discrete reasoning, we design a set of operations $\mathcal{OP}$ as shown in Table~\ref{Operators}.
The set contains 11 operations and each one represents a reasoning unit.
Specifically, for questions that need to be answered by calculation, we design operations $\texttt{ADDITION/DIFF}$ to represent addition and subtraction.
For questions which need to be answered by counting or sorting, we also design operations $\texttt{COUNT}$, $\texttt{MAX/MIN}$, $\texttt{ARGMAX/ARGMIN}$, and $\texttt{ARGMORE/ARGLESS}$.
The rest operations $\texttt{KEY\_VALUE}$ and $\texttt{SPAN}$ are used to extract spans from the question and the passage.
To incorporate them into OPERA, each operation is denoted as a tuple.
Formally, $i$-th operation $\mathcal{OP}_i$ is $\langle\mathbf{E}^{OP}_i, f_{i}(\cdot)\rangle$, where $i\in \{1,2,...,n\}$ and $n$ is the numbers of operations.
$\mathbf{E}^{OP}_i \in \mathbb{R}^{d_h}$ represents the learnable embedding of the $i$-th operation. $f_i(\cdot)$ is a neural executor parameterized with trainable matrices $\mathbf{W}^{OP}_{q,i}$, $\mathbf{W}^{OP}_{k,i}$ and $\mathbf{W}^{OP}_{v,i} \in \mathbb{R}^{d_h \times d_h}$. The neural executor $f_i(\cdot)$ is capable of performing execution of $\mathcal{OP}_i$ on the given context.
Specifically, it takes the representation of context as input and outputs the executed representation as $\mathbf{m}^{OP}_i$ (\cref{sec:operation-pivoted Reasoning Module}).

\label{sec:OPERA}

\subsection{Architecture of OPERA}

\subsubsection{Context Encoder}
\label{sec:context_encoder}
The context encoder aims to learn the contextual representation of the input.
Formally, given a question $Q$ and a passage $P$, we concatenate them into a sequence and feed it into a pre-trained language model~\cite{liu2019roberta,clark2020electra,lan2019albert} to obtain their whole representation $\mathbf{H} \in \mathbb{R}^{l\times d_h}$.
After that, we split $\mathbf{H}$ into the question and passage representations, 
which are respectively denoted as $\mathbf{H}^Q \in \mathbb{R}^{l_{q}\times d_{h}}$ and $\mathbf{H}^P\in \mathbb{R}^{l_{p}\times d_{h}}$. $l_q$, $l_p$, and $l$ are the number of tokens in question, passage and concatenation of them. $d_{h}$ is the dimension of the representations.

\subsubsection{Operation-pivoted Discrete Reasoning}
\label{sec:operation-pivoted Reasoning Module}
The operation-pivoted reasoning module is composed of an operation selector and an operation executor.
The operation selector is adopted to select operations related to the given question.
The operation executor is responsible for imitating the execution of the selected operations with an attention mechanism.

\paragraph{{Operation Selector}}

To imitate discrete reasoning, existing methods usually adopt a logical form generated by a semantic parser to address this task.
However, these methods suffer severely from the cascade error, where a little perturbation in the logical form may result in wrong answers.
Therefore, we propose to map each question into an operation set, instead of logical forms.
Namely, we intend to select relevant operations from the $\mathcal{OP}$.
To this end, we adopt a bilinear function to compute the similarity between each operation and the question and normalize them with a softmax as follows:
\vspace{-0.1cm}
\begin{equation}
  p(O|Q,P) = \mathtt{softmax}(\mathbf{E}^{OP} \mathbf{W} \mathbf{h}^{Q}),
\vspace{-0.1cm}
\end{equation}
where $\mathbf{E}^{OP} \in \mathbb{R}^{n\times d_h}$ is a learnable parameter, which demotes the operation embedding matrix.
$\mathbf{h}^{Q} \in \mathbb{R}^{d_h}$ is the representation of the question, which is obtained by executing weighted pooling on the $\mathbf{H}^{Q}$.
$\mathbf{W}\in \mathbb{R}^{d_h\times d_h}$ is a parameter matrix and $p(O|Q,P)$ is the
distribution over operations. 
  
\paragraph{{Operation Executor}}
The operation executor is responsible for performing the execution of the selected operations over the given context. 
Inspired by previous studies \cite{andreas2016neural,gupta2019neural}, we implement the operation executor based on the neural module network, which takes advantage of neural network in fitting and generalization, and the composition characteristics of symbolic processing. 
Specifically, for each operation $\mathcal{OP}_i=\langle\mathbf{E}^{OP}_i, f_{i}(\cdot)\rangle$, $i=\{1,2,...,n\}$, we use a multi-head cross attention mechanism \cite{vaswani2017attention} to implement $f_{i}(\cdot)$.
In detail, we leverage the embedding of each operation $\mathbf{E}^{OP}_i$ as query and the representations of the whole input sequence $\mathbf{H}$ as keys and values, respectively, to model $f_{i}(\cdot)$ as follows:
\vspace{-0.1cm}
\begin{align}
    \mathbf{\alpha}^{OP}_{i}&=\mathtt{softmax}(\frac{(\mathbf{E}^{OP}_{i}\mathbf{W}^{OP}_{q,i})(\mathbf{H}\mathbf{W}^{OP}_{k,i})^{T}}{\sqrt{d_h}}), \\
    \mathbf{m}^{OP}_{i} &= \mathbf{\alpha}^{OP}_{i}(\mathbf{H}\mathbf{W}^{OP}_{v,i}),
\vspace{-0.1cm}
\end{align}
where $\mathbf{W}^{OP}_{q, i}, \mathbf{W}^{OP}_{k, i},\mathbf{W}^{OP}_{v, i} \in \mathbb{R}^{d_h\times d_h}$ are the parameter matrices in executor of operation $\mathcal{OP}_i$. 
$\mathbf{m}^{OP}_i \in \mathbb{R}^{d_h}$ is the 
representation of the execution result of the $i$-th operation.

Finally, we softly integrate all of the execution results as the final output $\mathbf{h}^{OP}\in \mathbb{R}^{d_h}$ with the distribution $p(O|Q,P)$ as follows:
\vspace{-0.1cm}
\begin{equation}
\label{eq:h_op}
        \mathbf{h}^{OP} = \sum_{i=1}^{n}p(O=\mathcal{OP}_i|Q,P)\mathbf{m}^{OP}_{i}.
\vspace{-0.1cm}
\end{equation}
The operation-aware semantic representation $\mathbf{h}^{OP}$ is further fed into the prediction module to reason the final answer (\cref{sec:prediction_module}).

As described above, OPERA introduces operations that assist in understanding questions and integrates them into the model to perform discrete reasoning.
Therefore, it achieves an advantage in the reasoning capability and interpretability over the previous multi-predictor-based methods~\cite{hu2019multi,chen2020question,zhou2021evidr}.
Moreover, soft execution and composition of operations in OPERA alleviate the cascaded error that the semantic parsing methods \cite{ran2019numnet,chen2019neural} suffer from.
More experiments and analyses about reasoning ability and interpretability are illustrated in \cref{sec:main_results} and \cref{sec: Interpretability_Analysis}.

\subsubsection{Prediction Module} 
\label{sec:prediction_module}
In this section, we introduce the prediction module to derive different types of answers via multi-predictors.
Each predictor first reasons out a derivation and then performs execution to obtain the final answer.
This answer prediction procedure is formalized as follows:
\begin{equation}
\label{eq:answer_predict}
p(A|Q,P,O) =\!\sum_{D\in\mathcal{D}}{\mathbb{I}(g(D)=A)p(D|Q,P,O)},
\end{equation}
where $\mathbb{I}(g(D)=A)$ is an indicator function with value 1 if the answer $A$ can be derived from a derivation executor $g(\cdot)$ based on $D$, and 0 otherwise. $p(D|Q,P,O)$ models the derivation prediction.
Specifically, a derivation $D=\langle T, L \rangle$ includes an answer type $T$ and a corresponding label $L$.
For example, in Table~\ref{translation}, the textual answer $A$ of the question ``\textit{how many yards was the longest field goals in the game}” is ``$80$”. The possible derivations $\mathcal{D}$ to this answer include a span $\langle \mathtt{Span}, (100,102) \rangle$, and an arithmetic expression $\langle \mathtt{AE}$, $(0*29)+(1*80) \rangle$.
Inspired by previous studies \cite{chen2020question,zhou2021evidr}, the derivation predictor 
\vspace{-0.1cm}
\begin{equation}
p(D|Q,\!P,\!O)\!=\!\sum_{T\in\mathcal{T}}p_{T}(L|Q,\!P,\!O)p(T|Q,\!P,\!O)
\vspace{-0.1cm}
\end{equation}
is decomposed into an answer type predictor $p(T|Q,P,O)$ and corresponding label predictors 
$p_{T\in\mathcal{T}}(L|Q,P,O)$ where $\mathcal{T}\!=\!\{$\textit{Question Span}, \textit{Passage Span}, \textit{Count}, \textit{Arithmetic Expression}, \textit{Multi-spans}$\}$ includes all the answer types defined in this paper. 
Each label predictor takes question-passage representation $\mathbf{H}$ and the operation-pivo representation $\mathbf{h}^{OP}$ as input and calculates the probability of label $L$.
Specifically, these label predictors are specified as follows and more details are shown in Appendix \ref{sec:model_arichitecture}.

\noindent{\textbf{Question / Passage Span}} The probability of a question/passage span is the product of the probabilities of the start index and the end index.
Following MTMSN \cite{hu2019multi}, we use a question-aware decoding strategy to predict the start and end index across the input sequence,  respectively.

\noindent{\textbf{Count}} As indicated in QDGAT \cite{chen2020question}, questions with 0-9 as answers account for 97\% in all the count questions. Hence, such questions are modeled as a 10-class (0-9) classification problem.

\noindent{\textbf{Arithmetic Expression}} Similar to NAQANet \cite{dua2019drop}, we first assign a sign (positive, negative, or zero) to each number in the context and then compute the answer by summing them.

\noindent{\textbf{Multi-spans}} Inspired by \citet{segal2019simple}, the multi-span answer (a set of non-contiguous spans) is derived with a sequence labeling method, in which each token of the input is tagged with $\mathtt{BIO}$ labels.
Finally, each span which is tagged with continuous $\mathtt{B}$ and $\mathtt{I}$ is taken as a candidate span.

\label{sec:training and inference}

\subsection{Learning with Weak Supervision}
\subsubsection{Training Instance Construction}
\label{sec:Heuristic Rules}
Each training instance is originally composed of a passage $P$, a question $Q$, and answer text $A$.
Since the derivations (i.e., labels for the spans, arithmetic expressions, and count) are not provided, weak supervision is adopted in OPERA.
Specifically, for each training instance, given the golden textual answer $A$, we heuristically search all the possible derivations $\mathcal{D}$ as supervision signals, each of which can derive the correct answer $A$.
Table~\ref{translation} shows an example of $\mathcal{D}$.

In addition, we propose heuristic rules to map a question to its related operations denoted as $\mathcal{O} \subseteq \mathcal{OP}$. 
For example, to detect the operations intimated by the question $Q$ in Table~\ref{translation}, we design a question template ``\textit{how many yards [Slot] longest [Slot]}” which maps matched questions to the operation $\texttt{MAX}$. 
Overall, a training instance can be constructed as a tuple $\langle P,Q,A,\mathcal{O},\mathcal{D}\rangle$.
The one-shot heuristic rules to obtain operation labels reduce the cost of human annotations. Moreover, when applying OPERA to other discrete reasoning MRC tasks, both the operations $\mathcal{OP}$ and the heuristic rules can be extended and adjusted if necessary. Fortunately, there is no need to construct strict logical forms in our architecture, but only the set of lightweight operations involved in the question. 
It tremendously reduces the difficulty of adapting OPERA to other discrete reasoning MRC tasks.

Meanwhile, we analyze the distribution of operations in the training set.  
More details about the heuristic rules for mapping questions to operations and the operation distribution in the dataset are respectively given in the Appendix~\ref{appendix:rules} and~\ref{appendix:dis of operations}.

\begin{table}[t]
  \small
  \centering
	\begin{tabularx}{0.48\textwidth}{p{0.24in} p{2.47in}}
		\hline \hline
        $P$ & \textit{...Oakland would take the lead in the third quarter with wide 
            receiver Johnnie Lee Higgins catching a \textcolor{cyan}{29}-yard touchdown 
            pass from Russell, followed up by an \textcolor{red}{80}-yard punt return for a touchdown ...}\\
       $Q$ & \textcolor{blue}{\textit{How many yards was the longest field goals}}\\
       $A$ & $80$ \\
      $\mathcal{O}$ & $\texttt{MAX}$~$\Leftarrow$~\textcolor{blue}{\textit{How many yards [Slot] longest [Slot]}}\\
      $\mathcal{D}$ & $\langle\mathtt{Span},\textcolor{red}{(100, 102)}\rangle$;~~ $\langle \mathtt{AE},(0*\textcolor{cyan}{29})+(1*\textcolor{red}{80})\rangle$\\
      \hline \hline
	\end{tabularx} 
	\caption{\label{translation} An example of building training instances.}
\end{table} 

\subsubsection{Joint Training}
The training objective consists of two parts, including the loss for answer prediction and operation selection.
The loss for answer prediction $\mathcal{L}_{a}$ is 
\vspace{-0.1cm}
\begin{equation}
\begin{aligned}
    \mathcal{L}_{a} &= -{{\log{p(A|Q,P)}}}.\\
\end{aligned}
\label{eq6}
\vspace{-0.1cm}
\end{equation}
Note that the calculation of loss $\mathcal{L}_{a}$ takes all possible derivations that can obtain the correct answer $A$ into account, which means that OPERA does not require labeling answer types for training. 
In addition, to learn better alignment from a question to operations,
we introduce auxiliary supervision for the operation selector and calculate the loss 
\vspace{-0.1cm}
\begin{equation}
     \mathcal{L}_{op} = -\sum_{O \in \mathcal{O}}{\log{ p(O|Q,P)}},
\vspace{-0.1cm}
\end{equation}
where $\mathcal{O}$ indicates the operations provided by the heuristic rules. 
Finally, OPERA is optimized by minimizing the loss 
$\mathcal{L} = \mathcal{L}_{a} + \lambda \mathcal{L}_{op}$
where $\lambda$ is a hyperparameter as a trade-off of the two objectives.

\section{Experiment}
\subsection{Dataset and Evaluation}
We conduct experiments on the following two MRC datasets to examine the discrete reasoning capability of our model.
We employ Exact Match (EM) and F1 score as the evaluation metrics.

\paragraph{DROP} 
Question-answer pairs in DROP dataset \cite{dua2019drop} are crowd-sourced based on passages collected from Wikipedia. 
In detail, it contains 96.6K question-answer pairs, where 77400/9536/9615 samples are for training/development/test. 
Three kinds of answers are involved in the raw dataset, i.e., NUMBER (60.69\%), SPANS (37.72\%), and DATE (1.59\%). 

\paragraph{RACENum}
To investigate the generalization capability of OPERA, we compare OPERA to other strong baselines on samples of RACE \cite{lai2017race}. 
Following \citet{chen2020question}, we sample instances from RACE, denoted as RACENum, where the question of each instance starts with ``how many”.  
To conveniently evaluate the models on RACENum, we convert the format of instances in RACENum into the same as DROP, since RACE is a multi-choice MRC dataset.
RACENum is divided into two categories, i.e., middle/high school exam (RACENum-M/H). They respectively contain 609 and 565 questions, where the scale is a bit different from that reported in \citet{chen2020question}\footnote{Since no released code for dataset construction, we implement it referred to the algorithm in \citet{chen2020question}.}.

\subsection{Baselines}
We compare OPERA with various prior systems in terms of reasoning capability and interpretability.

\noindent \textbf{w/o Pre-trained Language Model:}
NAQANET \cite{dua2019drop} leverages several answer predictors to produce corresponding types of answers, including a span, count, and arithmetic expression. NumNet~\cite{ran2019numnet} leverages a graph convolution network to reason over a number graph aggregated relative magnitude among numbers. 

\noindent \textbf{w/ Pre-trained Language Model:}
GenBERT \cite{geva2020injecting} injects reasoning capability into BERT by pre-training with large-scale numerical data.
Based on NAQANET, MTMSN \cite{hu2019multi} adds a negation predictor to solve the negative question and re-rank arithmetic expression candidates. 
NeRd \cite{chen2019neural} is essentially a generative semantic parser that maps questions and passages into executable logical forms. 
ALBERT-Calc was proposed  for DROP by combining ALBERT with several predefined answer predictors \cite{andor2019giving}. 
NumNet+ employs a pre-trained model to further boost the performance of NumNet.
QDGAT \cite{chen2020question} builds a heterogeneous graph composed of entity and value nodes upon RoBERTa and utilizes a question-directed graph attention network to reason over the graph.
EviDR \cite{zhou2021evidr}, an evidence-emphasized MRC model, performs reasoning over a multi-grained evidence graph based on ELECTRA.

\subsection{Implementation Details}
\label{appendix:details}
We utilize adam optimizer \cite{kingma2015adam} with a cosine warmup mechanism and set the weight of loss $\lambda=0.3$ to train the model. The hyper-parameters are listed in Table \ref{Hyperparameters_OPERA}, where BLR, LR, BWD, WD, BS, and $d_h$ respectively represent the learning rate of the encoder, the learning rate of other parts of the model, the weight decay of the encoder, the weight decay of other parts of the model, batch size and hidden size of the model. Each operation is neutralized with a multi-head attention layer with $n_h$ heads and $d_h$ dimension.

\begin{table}[t]
  \small
  \centering
  \resizebox{0.48\textwidth}{!}{
	\begin{tabular}{lcccccccc}
		\toprule
        & BLR  & LR & BWD & WD &Epochs & BS & $n_h$ &$d_h$ \\
        \midrule
        RoBERTa &1.5e-5 & 5e-4 & 0.01& 5e-5& 12&16 & 16 & 1024\\
        ELECTRA &1.5e-5 & 5e-4 & 0.01& 5e-5& 12&16 & 16 & 1024\\
        ALBERT  &3e-5 & 1e-4 & 0.01& 5e-5& 8 &128 & 64 & 4096\\
		\bottomrule
	\end{tabular} }
	\caption{\label{Hyperparameters_OPERA} Hyperparameters settings for training OPERA.}
\end{table}

\subsection{Main Results}
\label{sec:main_results}
\subsubsection{Results on DROP and Analysis}

Table \ref{r|esults} shows the overall results of OPERA and all the baselines on the DROP dataset.
OPERA achieves comparable and even higher performance than the recently available methods. 
Specifically, OPERA(RoBERTa) achieves comparable performance to QDGAT with advantages of 0.32 EM and 0.42 F1. OPERA(ELECTRA) exceeds EviDR by 0.89 EM and 0.90 F1 and OPERA(ALBERT) outperforms ALBERT-Calc by 4.84 EM and 4.24 F1.
Moreover, the voting strategy is employed to ensemble 7 OPERA(ALBERT) models with different random seeds, achieving 86.26 EM and 89.12 F1 scores. We think the better performance comes from the modeling of discrete reasoning over text via operations, which mines more semantic information of context and explicitly integrates them into the answer prediction.

\begin{table}[t]
  \small
  \centering
  \resizebox{0.49 \textwidth}{!}{
  \setlength{\tabcolsep}{1mm}{
	\begin{tabular}{l  cccc}
		\toprule
		\multirow{2}*{\bf Method}     &\multicolumn{2}{c}{\bf Dev } &   \multicolumn{2}{c}{\bf Test}\\ 
		 \cmidrule(lr){2-3} \cmidrule(lr){4-5}
		&\bf EM &\bf F1  &  \bf EM & \bf F1 \\ 
		\midrule
		\textbf{w/o Pre-trained Models}\\
		NAQANet  & 46.20 &49.24 &44.07 &47.01\\
		NumNet  & 64.92 &68.31 &64.56 &67.97\\
		\midrule
		\textbf{w/ Pre-trained Models} \\
	    GenBERT  & 68.80 & 72.30&68.6 &72.35 \\
	    MTMSN  &76.68 & 80.54&  75.88& 79.99\\
    	NeRd &  78.55 & 81.85 & 78.33 & 81.71\\ 
	    ALBERT-Calc  &80.22 &83.98 &79.85 &83.56\\
        NumNet+   &81.07&84.42 & 81.52& 84.84\\
        EviDR & 82.09&85.14 &82.55 &85.80 \\
        QDGAT & 82.74 & 85.85 & 83.23 & 86.38 \\
		\midrule
        \textbf{Single Model Results} \\
        OPERA(RoBERTa)   &{83.74} &{86.52} &{83.55}&{86.80}\\ %
        OPERA(ELECTRA)   &{83.86} &{86.66} &{83.46}&{86.70}\\ %
        OPERA(ALBERT)   &\bf{84.86} & \bf{87.54} & \bf{84.69}&\bf{87.80} \\  %
        \midrule
        \textbf{Ensemble Results} \\
		OPERA  &{ 86.79}&{ 89.41}&{ 86.26}&{ 89.12}\\ %
		\midrule
		Human  & & & 94.90& 96.42\\
		\bottomrule
	\end{tabular}
	} }
	\caption{\label{r|esults} Results on the DROP dataset.
	We solely compare with QDGAT, but leaving QDGAT$_{p}$ alone, since we focus on the reasoning mechanism in this work, while QDGAT$_{p}$ is a variant of QDGAT with data augmentation ~\cite{chen2020question}.
	}
\end{table}

\subsubsection{Results on RACENum}
To investigate the generalization of OPERA for discrete reasoning, we additionally compare OPERA with QDGAT and NumNet+ on the RACENum dataset.
We directly evaluate the three models without fine-tuning on RACENum due to its small scale.
As Table \ref{RACENum} shows, the scores of models on the RACENum dataset are generally lower than that on the DROP dataset, which is attributed to the lack of in-domain training data.
Nevertheless, the performance of OPERA significantly outperforms NumNet+ and QDGAT by a large margin of more than 3.49 EM and 3.53 F1 score on average.
It indicates that OPERA has better generalization ability.

\subsection{Interpretability Analysis}
\label{sec: Interpretability_Analysis}
Interpretability is an essential property for evaluating an MRC model.
We analyze the interpretability of OPERA from the following two stages: (1) mapping from questions to operations, and (2) mapping from operations to answers.

\paragraph{Mapping from Question to Operation}
\begin{table}[t]
  \small
  \centering
  \resizebox{0.48\textwidth}{!}{
  \setlength{\tabcolsep}{1.2mm}{
	\begin{tabular}{l cccc cc}
	    \toprule
		\multirow{2}*{\bf Method}  &  \multicolumn{2}{c}{\bf RACENum-M} &   \multicolumn{2}{c}{\bf\bf RACENum-H} &   \multicolumn{2}{c}{\textbf{Avg}}\\
		\cmidrule{2-3} \cmidrule{4-5} \cmidrule{6-7}
         &  \bf{EM} & \bf{F1}       &  \bf{EM} & \bf{F1} &  \bf{EM} & \bf{F1} \\
		\midrule
	    NumNet+ & 41.71&41.82 & 29.73&29.73 &35.94&36.00\\
		QDGAT   & {44.01} & {44.01} & {28.85} & {28.85}&36.71&36.71 \\
		OPERA    & \bf{47.62}& \bf{47.62} & \bf{32.21}& \bf{32.30}&\bf{40.20}&\bf{40.24}\\
		\bottomrule
	\end{tabular} }}
	\caption{\label{RACENum} The performance of RoBERTa-based models on the RACENum dataset without finetuning.}
\end{table}

\begin{figure}[t]
  \centering
	\includegraphics[width=2.8in]{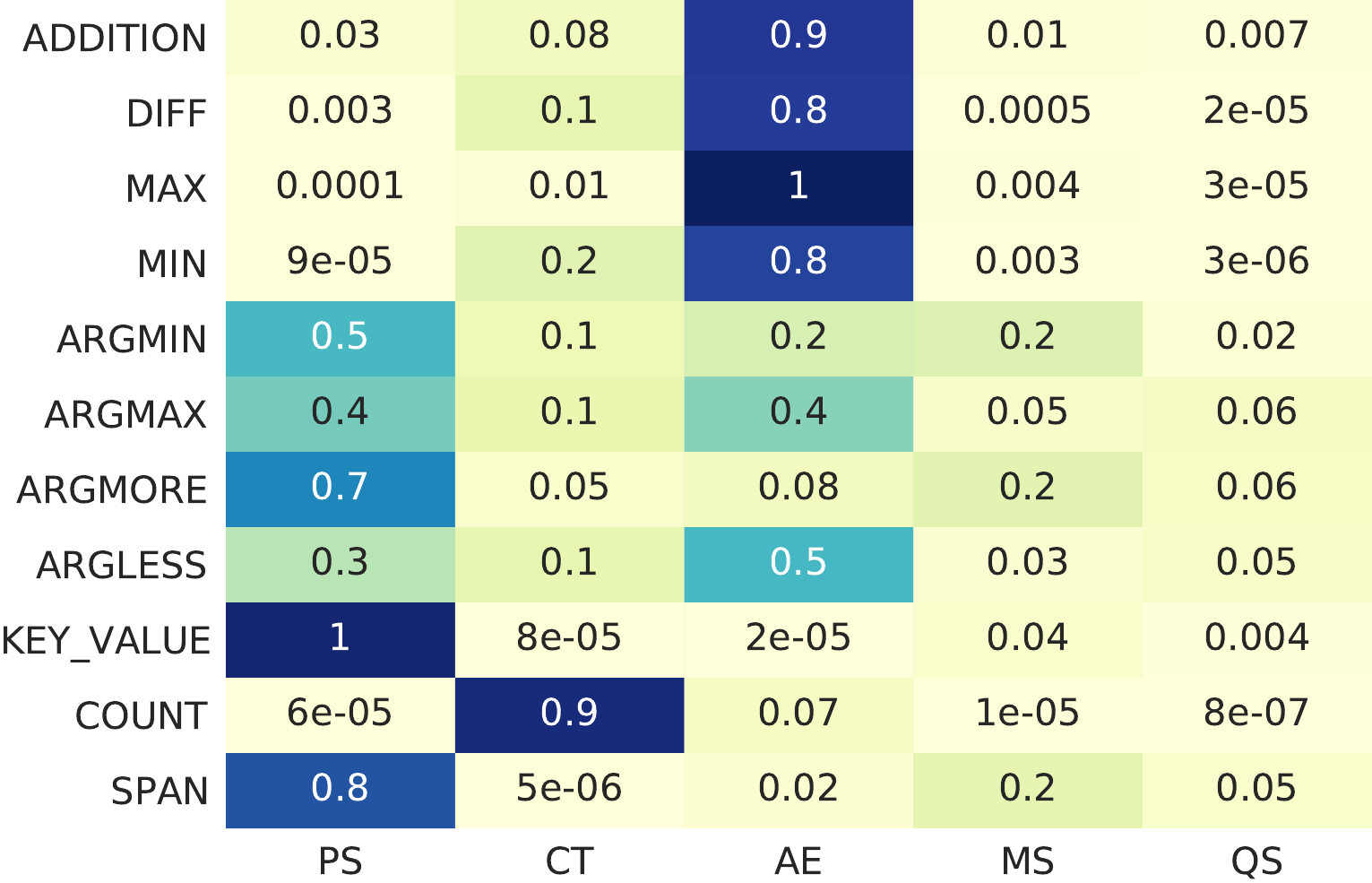}
	\caption{\label{op_answer_type}Statistical correlations between operations (the vertical axis) and answer types (the horizontal axis).
	PS, CT, AE, MS and QS respectively indicate
	\textit{passage span}, \textit{count}, \textit{arithmetic expression}, \textit{multi-spans} and \textit{question span}.
	}
\end{figure}

To explicitly show the correlations between questions and related operations, we manually evaluate the performance of the operation selection on 50 samples on the development set of DROP.
Specifically, precision@n (P@n) is used as the evaluation metric, i.e., judging whether the top $n$ predicted operations contain the correct ones according to questions. 
We finally achieve 0.88 on P@1 and 0.98 on P@2 for our model OPERA, which indicates that the operation selection module can accurately predict interpretable operations.

\paragraph{Mapping from Operation to Answer}
We explore the relations between operations and final answer types based on model predictions on the development set of DROP.
Specifically, for each type of answer, the predicted operation distributions are summed over all samples and normalized, which derives a relation matrix as shown in Figure \ref{op_answer_type}.
We can observe that obvious correlations between operations and answer types exist.
\texttt{ADDITION}, \texttt{DIFF}, \texttt{MAX} and \texttt{MIN} obviously correspond to \textit{Arithmetic Expression}. 
The top 3 answer types for \texttt{KEY\_VALUE} and \texttt{SPAN} are \textit{Passage Span}, \textit{Multiple Spans}, and \textit{Question Span}.
\texttt{COUNT} probably maps to \textit{Count} answer type.
\texttt{ARGMORE}, \texttt{ARGLESS}, \texttt{ARGMAX}, and \texttt{ARGMIN} are usually used for both \textit{Passage Span} and \textit{Arithmetic Expression}.

\subsection{Ablation Study}

\begin{figure}[t]
  \centering
	\includegraphics[width=2.9in]{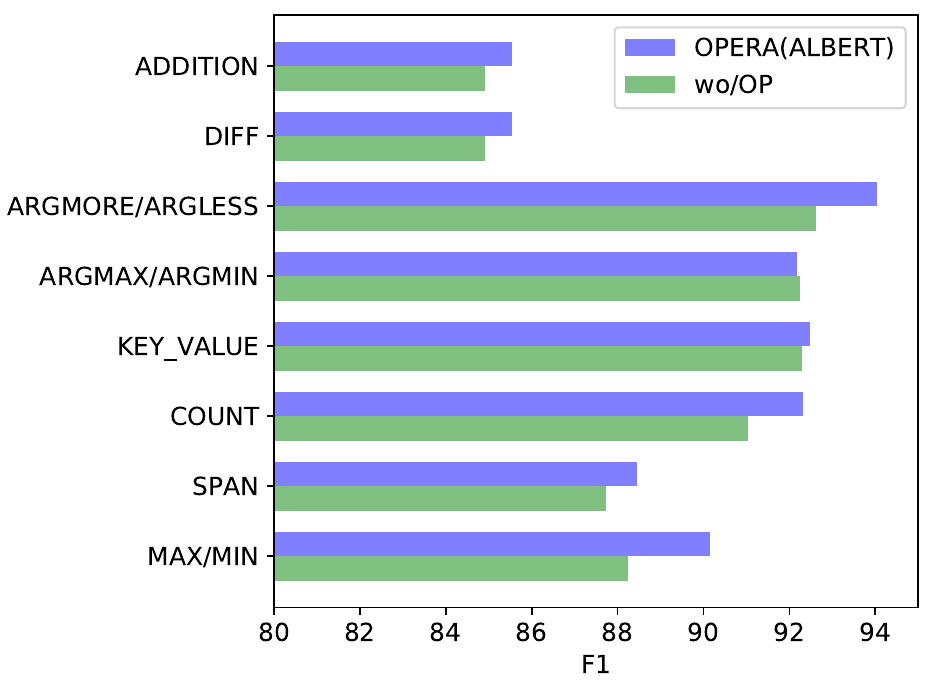}
	\caption{\label{performance_with_op} Ablation study on all the subsets of DROP containing some specific operations.
	}
\end{figure}

\paragraph{Effect of Operations for OPERA} 
As shown in Table \ref{Diff_answer_type}, we first remove the loss of operation selection (w/o $\mathcal{L}_{op}$) by setting $\lambda=0$, resulting in the performance degradation of 0.74 EM / 0.52 F1 points and 0.14 EM / 0.18 F1 points for OPERA based on RoBERTa and ALBERT, respectively. It indicates that the supervision for explicitly learning the alignment from a question to operations contributes to the reasoning capability of OPERA. 
Yet our approach is somewhat limited by the fact that the operation selector needs an auxiliary training task to work better, and heuristics rules are required to map questions into an operation set as training labels.
Furthermore, we remove the total operation-pivoted reasoning module (w/o OP), the performance respectively declines by 1.06 EM / 0.79 F1 points and 0.85 EM / 0.75 F1 points for OPERA(RoBERTa) and  OPERA(ALBERT). 
We also conduct the ablation study on the subsets containing a specific operation. As shown in Figure \ref{performance_with_op}, OPERA achieves better performance than OPERA w/o OP on the majority of subsets.
Overall, it confirms that integrating the operation-pivoted discrete reasoning mechanism contributes to the reasoning ability of the model.

\begin{table}[t]
  \small
  \centering
  \resizebox{0.48 \textwidth}{!}{
  \setlength{\tabcolsep}{1mm}{
	\begin{tabular}{lccc}
	    \toprule
		\multirow{2}*{\bf Method}  & \multicolumn{1}{c}{\bf NUM (60.69\%)}  & \multicolumn{1}{c}{\bf SPANS (37.72\%)} & \multicolumn{1}{c}{\bf Overall}\\ 
	   \cmidrule(lr){2-2} \cmidrule(lr){3-3} \cmidrule(lr){4-4} 
		&\bf EM / F1  &  \bf EM / F1  &  \bf EM / F1 \\ 
    	\midrule
        OPERA(RB) &        \bf{85.91 / 86.14}    & \bf{83.89 / 88.90}   &  \bf{83.74 / 86.52} \\  %
         w/o  $\mathcal{L}_{op}$ &  85.83 / 86.00 &    82.68 / 88.02 &           83.00 / 86.00 \\ %
         w/o OP &  85.36 / 85.53    &      82.53 / 87.06   &  82.68 / 85.73\\ %
        \midrule
         OPERA(AB) & \bf{86.39 / 86.58}     & 86.30 /  \bf{90.99}   &  \bf{84.86 / 87.54} \\  %
         w/o  $\mathcal{L}_{op}$  & 86.08 / 86.24 &  \textbf{86.39} / 90.96 & 84.72 / 87.36\\ %
         w/o OP & 85.62 / 85.89   &   84.95 / 89.83   &  84.01 / 86.79 \\ %
		\bottomrule
	\end{tabular} } }
	\caption{\label{Diff_answer_type} Ablation study on the dev set of DROP. RB and AB mean RoBERTa and ALBERT, respectively.}
	\label{ex:ablation}
\end{table}

\begin{table*}[t]
  \scriptsize
  \centering
  \resizebox{1\textwidth}{!}{
  \setlength{\tabcolsep}{2mm}{
	\begin{tabular}{m{1.4in} m{2.0in} m{0.5in} m{0.5in} p{0.85in}}
	    \toprule
		\bf {Question-Answer} & \bf {Passage} & \bf {NumNet+} & \bf{QDGAT} & \bf{OPERA} \\ 
    	\midrule
    {\textbf{Q}: \textcolor{orange}{How many total yards} of touchdown passes were there?  \;\;\;\;\;\;\;\;\;\;\;\;\;\;\;\;\;\;\;\;\;\;\;\;\;\; \textbf{A}: 73} & {...  
    receiver Johnny Knox on a \textcolor{blue}{23}-yard touchdown pass. Afterwards, the Falcons took the lead as quarterback Matt Ryan completed a \textcolor{blue}{40}-yard touchdown pass to wide receiver Roddy White and a \textcolor{blue}{10}-yard touchdown pass to tight end Tony Gonzalez
    ...} & \tabincell{l}{\textbf{AnswerType}: \\ Count \\ \textbf{Answer}: 0 } & \tabincell{l}{\textbf{AnswerType}: \\Count \\ \textbf{Answer}: 0}& \tabincell{l}{\textbf{Top-1 OP}: \textcolor{orange}{\texttt{ADDITION}} \\ \textbf{AnswerType}: \\Arithmetic Expression  \\  \textbf{Answer}: \\23+40+10=73 }\\
    	\midrule
    {\textbf{Q}: \textcolor{orange}{Which period} was Wolf executive and player personnel director with the Oakland Raiders \textcolor{orange}{longer for}, 1963-1974 or 1979-1989? \;\;\;\;\;\;\;\;\;\;\;\;\;\;\;\;\;\;\;\;\;\;\;\;\; \textbf{A}: 1963-1974} &
    Wolf only had a brief stint with the Jets between 1990 and 1991, while most of his major contributions occurred as an executive and player personnel director with the Oakland Raiders (\textcolor{blue}{1963-1974}, \textcolor{blue}{1979-1989}), and later as General Manager...
    &  \tabincell{l}{\textbf{AnswerType}: \\Passage Span  \\ \textbf{Answer}: \\1979-1989 }& \tabincell{l}{\textbf{AnswerType}: \\Passage Span \\ \textbf{Answer}: \\1979-1989} & \tabincell{l}{\textbf{Top-1 OP}: \textcolor{orange}{\texttt{SPAN}} \\ \textbf{AnswerType}: \\Passage Span  \\  \textbf{Answer}: \\1963-1974 } \\
    
		\bottomrule
	\end{tabular} }
	}
	\caption{\label{case_study} The cases from the development set of DROP. The predictions from the state-of-the-art model NumNet+ and QDGAT are shown. The last column indicates our predicted answers and Top-1 operations.}
\end{table*}

\paragraph{Probe on Answer Types and Language Models} 
As reported in Table \ref{Diff_answer_type}, we observe that the performance on the NUMBER(NUM) and SPANS questions, which together account for 98.4\% of the total, respectively declines by 0.55 EM / 0.61 F1 and 1.36 EM / 1.84 F1 when removing operation-pivoted reasoning mechanism from OPERA(RB). It demonstrates that this mechanism contributes to various answer types.
Also, we respectively evaluate the performance of OPERA based on RoBERTa and ALBERT. We observe that OPERA(ALBERT) outperforms OPERA(RoBERTa) due to the stronger capability of semantic representation. Furthermore, integrating this mechanism consistently contributes to the performance of OPERA no matter it is based on RoBERTa or ALBERT. It indicates that OPERA could compensate for the discrete reasoning capability of language models.

\subsection{Case Study}
We show two examples from the development set of DROP to illustrate the effectiveness of our model by comparing the results of different models in Table \ref{case_study}.
The first example shows that operation is essential for the prediction of answer type. 
NumNet+ and QDGAT fail to predict the correct answer since the answer type of “how many” questions are wrongly predicted to \textit{Count}. In contrast, OPERA can capture the \texttt{ADDITION} operation, which prompts the model to answer it with an \textit{arithmetic expression} predictor. 
The second example shows that OPERA has stronger reasoning capability. In the example, though NumNet+ and QDGAT correctly predict the answer type, the final answer is wrong. OPERA can utilize more semantic information for answer prediction with the help of the operation-pivoted discrete reasoning mechanism.

\section{Conclusion }
We propose a novel framework OPERA for machine reading comprehension requiring discrete reasoning.
Lightweight and one-shot operations and heuristic rules to map questions to an operation set are systematically designed.
OPERA can leverage the operations to enhance the model's reasoning capability and interpretability.
Experiments on DROP and RACENum demonstrate that OPERA achieves remarkable performance. 
Further visualization and analysis verify its interpretability.

\section{Acknowledge}
This work is supported by the project of the National Natural Science Foundation of China (No.U1908216) and the National Key Research and Development Program of China (No. 2020AAA0108600).

\bibliography{anthology}
\bibliographystyle{acl_natbib}

\clearpage
\appendix

\section{Appendix}
\subsection{Template-Based Heuristic Rules}
\label{appendix:rules}

\begin{table}[h]
  \small
  \centering
  \setlength\tabcolsep{10pt}
  \resizebox{0.48\textwidth}{!}{
\begin{tabular}{m{0.48\textwidth}}
 \toprule
 
 \bf Operations / Examples / Templates  \\ \midrule

\texttt{ADDITION/DIFF} \\ \textit{How many more yards was Kris Browns's first field goal over his second?} \\ How many [Slot] more/less [Slot] over [Slot]? \\ \midrule

\texttt{MAX/MIN} \\ \textit{How many yards was the longest field goal in the game?} \\ how many yards [Slot] longest/shortest [Slot]?\\ \midrule

\texttt{ARGMAX/ARGMIN } \\ \textit{Which player had the longest touchdown reception?} \\ Which player [Slot] longest/shortest [Slot]?\\
\midrule

\texttt{ARGMORE/ARGLESS} \\  \textit{Who scored more field goals, David Akers or John Potter?} \\ Who [Slot] more/less, [Slot] or [Slot]?\\ 
\midrule

\texttt{COUNT} \\  \textit{How many field goals did Kris Brown kick?} \\ How many field goals [slot]? \\\midrule

\texttt{KEY\_VALUE} \\ \textit{How many percent of Forth Worth households owned a car?} \\ How many percent of [Slot] \\ \midrule
 
\texttt{SPAN} \\  \textit{Which team scored the final TD of the game?} \\ Which team [Slot]\\

\bottomrule
\end{tabular} }
   \centering
   \caption{\label{Operators_patts} All the operations,  the corresponding examples and templates.}
\end{table}
In this section, we propose some general template-based heuristic rules to illustrate mapping from questions to operations.
For example, to detect the operations intimated by the question "\textit{how many yards was the longest field goals}" in Table~\ref{Operators_patts}, we design a question template $OP_{pat}$ "\textit{how many yards [Slot] longest/shortest [Slot]}" which maps matched questions to the operation $\texttt{MAX}$. 
Meanwhile, we humanly evaluate the quality of the heuristic rules.
Specifically, we sample 50 instances from the training set and ask three annotators to label the required operations for each question manually.
The final average F1 score of the three annotators is 86\%, which indicates that the heuristic rules can correctly predict most of the operations, while still 14\% to be noise for training.

\subsection{Distribution of the Operations}
\label{appendix:dis of operations}
\begin{figure}[ht]
  \centering
	\includegraphics[width=3in]{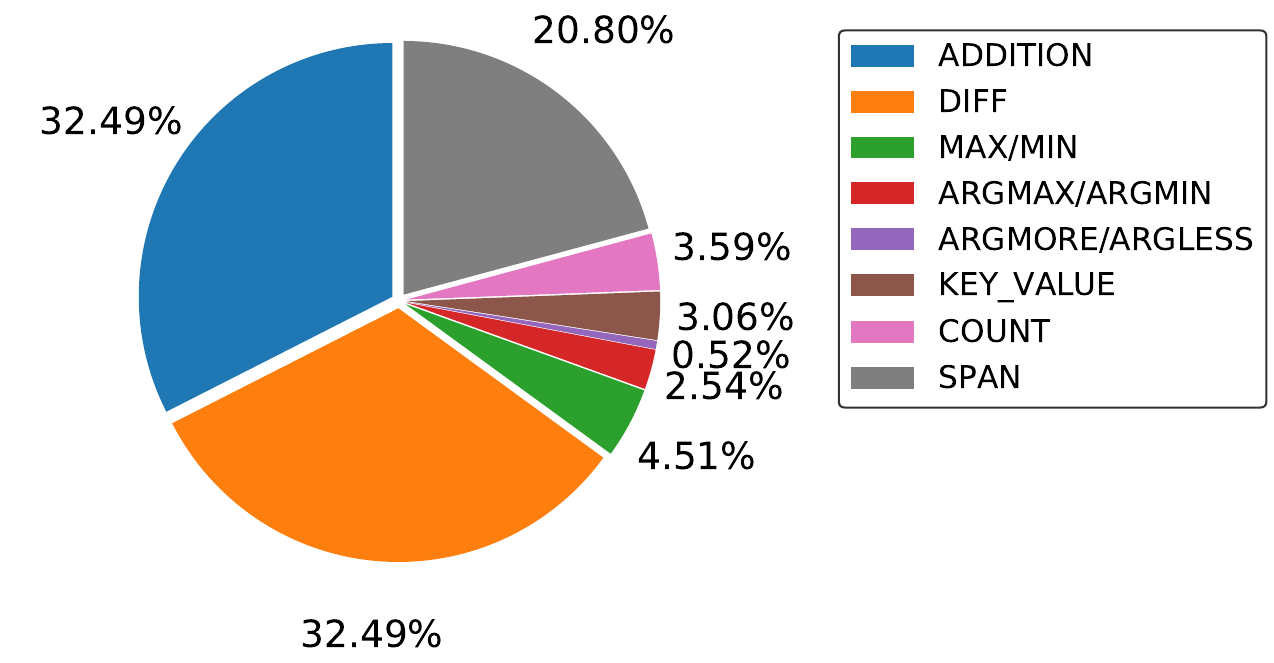}
	\caption{\label{op_statics} The Distribution of operations in the training set of DROP.}
\end{figure}

We analysis the distribution of operations in the training set of DROP, where \texttt{ADDITION}, \texttt{DIFF} and \texttt{SPAN} together accounts for more than 85\%. For other questions with span answers that requires sorting or comparison, some specific operations are involved, such as \texttt{ARGMAX} / \texttt{ARGMIN} / \texttt{ARGMORE} / \texttt{ARGLESS} and \texttt{KEY\_VALUE}.

\subsection{Details of Prediction Module}
\label{sec:model_arichitecture}
In this section, we reveal the architecture details of the prediction module, including a prediction module for answer type and five label predictors corresponding to different answer types.
$\mathtt{FFN}(\cdot)$ means a feed-forward network that consists of two linear projections with a GeLU activation function \cite{hendrycks2016bridging} and a layer normalization mechanism \cite{ba2016layer}.

\paragraph{Answer Type} 
The probability distribution of answer type choices $p(T|Q,P,O)$ is derived by a $\left |\mathcal{T}\right|$-classifier with $\mathbf{h}^Q$, $\mathbf{h}^P$ and $\mathbf{h}^E$ as input:
\begin{equation}
\begin{split}
    &\mathbf{h}^E =\sum_{O_i\in \mathcal{OP}}{p(O_i|Q,P)\mathbf{E}^{OP}_{i}},\\
    &p(T|Q,P,O) \propto \mathtt{FFN}( [\mathbf{h}^E;\mathbf{h}^{Q};\mathbf{h}^{P}]),
\end{split}
\end{equation}
where $\mathbf{h}^{Q}$ and $\mathbf{h}^{P} \in \mathbb{R}^{d_h}$ is the representation vector of question and passage calculated by weighted pooling with $\mathbf{H}^{Q}$ and $\mathbf{H}^{P}$, respectively. $\mathbf{E}^{OP}$ is the embedding matrix of operations.
    
\paragraph{Question/Passage Span}
Following MTMSN \cite{hu2019multi}, we use a question-aware decoding strategy to predict the start and end indices of the answer span. Specifically, we first compute a question representation vector $\mathbf{g}^{Q}$ via weighted pooling.
Then derive the probabilities of the start and end indices of the answer span denoted as $p_{s}$ and $p_{e}$:
\begin{equation}
    \begin{split}
    & \mathbf{M} = [\mathbf{h}^{OP};\mathbf{H};\mathbf{H} \odot \mathbf{g}^{Q}], \\
    & p_{s}, p_{e} \propto \mathtt{FFN}(\mathbf{M}),\\
    \end{split}
\end{equation}
where $\odot$ denotes element-wised product. $\mathbf{h}^{OP}$ is derived by Eq.~\ref{eq:h_op} and $\mathbf{H}$ is the representation of input sequence from context encoder. 

\paragraph{Count}
The count predictor is a 10-classifier with the operation-aware representation, all the mentioned number representation, question and passage representations as input. Specifically, when $N$ numbers exists, we gather the representation of all numbers $\mathbf{U} = (\mathbf{u}^{1}, \mathbf{u}^{2},..., \mathbf{u}^{N}) \in \mathbb{R}^{N\times d_h}$ from $\mathbf{H}$ and compute a global representation vector of numbers as $\mathbf{h}^{U}$.  Then compute the probability distribution of count answer $p_{c}$:
\begin{equation}
    \begin{split}
        \mathbf{\alpha}^{U} \propto \mathbf{U}\mathbf{W}^{U}, ~~\mathbf{h}^{U} = \mathbf{\alpha}^{U}\mathbf{U},\\
        p_{c} \propto \mathtt{FFN}([\mathbf{h}^{OP}; \mathbf{h}^{U}; \mathbf{h}^{Q};\mathbf{h}^{P}]),
    \end{split}
\end{equation}

\paragraph{Arithmetic Expression}
Similar to NAQANet \cite{dua2019drop}, we perform addition and subtraction over all the numbers mentioned in the context by assigning a sign (plus, minus, or zero) to each number. The probability $p_{sign}^{i}$ of the $i$-th number's sign is derived as below:
\begin{equation}
    \begin{split}
        p_{sign}^{i} \propto \mathtt{FFN}( [\mathbf{h}^{OP};\mathbf{u}^{i};\mathbf{h}^{Q};\mathbf{h}^{P}]).
    \end{split}
\end{equation}

\paragraph{Multi-Spans}
Inspired by \citet{segal2019simple}, the multi-span answer is derived with a sequence role labeling method over the input token sequence, denoted as $\mathtt{SRL}(\cdot)$. $p_{ms}$ is the probability distribution of token's \texttt{BIO} tag:
        \begin{equation}
        \begin{split}
            p_{ms} = \mathtt{SRL}([\mathbf{H};\mathbf{h}^{OP}]).
        \end{split}
    \end{equation}

\end{document}